\documentclass[letterpaper, 10 pt, conference]{ieeeconf}  

\IEEEoverridecommandlockouts                              

\usepackage[left=1.88cm,right=1.88cm,top=1.88cm,bottom=1.88cm]{geometry}
\usepackage{amsmath,amssymb,amsfonts}
\usepackage{cite}
\usepackage{graphicx}
\usepackage{textcomp}
\usepackage{xcolor}
\usepackage{bbm}
\let\labelindent\relax
\usepackage{enumitem}
\usepackage{tabularx}

\usepackage{times}
\usepackage{multicol}
\usepackage{soul}

\usepackage{graphicx}
\usepackage{bm}
\usepackage[nolist]{acronym}

\usepackage[linesnumbered,ruled]{algorithm2e}

\usepackage{mathtools}
\usepackage{array}
\usepackage{dblfloatfix}
\usepackage{subfigure}
\usepackage{diagbox}
\usepackage{etoolbox}\AtBeginEnvironment{algorithmic}{\small}

\newif\ifshowcomments
\showcommentsfalse 

\ifshowcomments
    \newcommand{\bae}[1]{\hl{[SB: #1]}\protect\color{black}} 
    \newcommand{\di}[1]{\hl{[DI: #1]}\protect\color{black}} 
    \newcommand{\ft}[1]{\hl{[FT: #1]}\protect\color{black}} 
    \newcommand{\jd}[1]{\hl{[JD: #1]}\protect\color{black}} 
    \newcommand{\ms}[1]{\hl{[MJ: #1]}\protect\color{black}} 
    \newcommand{\pg}[1]{\hl{[PG: #1]}\protect\color{black}} 
\else
    \newcommand{\bae}[1]{}
    \newcommand{\di}[1]{}
    \newcommand{\ft}[1]{}
    \newcommand{\jd}[1]{}
    \newcommand{\ms}[1]{}
    \newcommand{\pg}[1]{}
\fi

\usepackage{lipsum}
\usepackage[colorinlistoftodos]{todonotes}

\usepackage{amsthm}

\usepackage{bbm} 

\makeatletter
\let\NAT@parse\undefined
\makeatother
\usepackage{url}
\usepackage{xcolor}
\definecolor{mycitecolor}{RGB}{71, 191, 38}
\definecolor{mylinkcolor}{RGB}{40, 115, 201}

\usepackage{hyperref}
\hypersetup{
  colorlinks=true,
  citecolor=mycitecolor,
  linkcolor=mylinkcolor,
  urlcolor=mylinkcolor,
}

\newtheorem{problem}{Problem}

\usepackage{booktabs}
\usepackage{multirow}
\usepackage{arydshln}
\usepackage{subcaption}

\DeclareMathAlphabet{\mathmybb}{U}{bbold}{m}{n}

\allowdisplaybreaks

\title{\LARGE \bf
IANN-MPPI: Interaction-Aware Neural Network-Enhanced Model Predictive Path Integral Approach for Autonomous Driving
 }

\author{Kanghyun Ryu$^{1,2}$ \quad Minjun Sung$^{1,3}$ \quad Piyush Gupta$^1$ \quad Jovin D'sa$^1$ \\ Faizan M. Tariq$^1$   \quad David Isele$^1$ \quad Sangjae Bae$^1$
\thanks{
All work is done at HRI and paper is submitted and published while Kanghyun Ryu and Minjun Sung were employed by HRI. 
(Email: \texttt{kanghyun.ryu@berkeley.edu}, \texttt{mjsung2@illinois.edu}, \{\texttt{piyush\_gupta}, \texttt{jovin\_dsa}, \texttt{faizan\_tariq}, \texttt{disele}, \texttt{sbae}\} \texttt{@honda-ri.com}
$^1$Honda Research Institute, USA, San Jose, CA, 95134.  
$^2$University of California Berkeley, CA, USA, 94720. 
$^3$University of Illinois Urbana-Champaign, IL, USA, 61801.}
}

\begin{document}

\maketitle


\begin{abstract}
Motion planning for autonomous vehicles (AVs) in dense traffic is challenging, often leading to overly conservative behavior and unmet planning objectives. This challenge stems from the AVs' limited ability to anticipate and respond to the interactive behavior of surrounding agents. Traditional decoupled prediction and planning pipelines rely on non-interactive predictions that overlook the fact that agents often adapt their behavior in response to the AV’s actions. To address this, we propose Interaction-Aware Neural Network-Enhanced Model Predictive Path Integral (IANN-MPPI) control, which enables interactive trajectory planning by predicting how surrounding agents may react to each control sequence sampled by MPPI. To improve performance in structured lane environments, we introduce a spline-based prior for the MPPI sampling distribution, enabling efficient lane-changing behavior. We evaluate IANN-MPPI in a dense traffic merging scenario, demonstrating its ability to perform efficient merging maneuvers. Our project website is available at \href{https://sites.google.com/berkeley.edu/iann-mppi}{https://sites.google.com/berkeley.edu/iann-mppi}
\end{abstract}

\section{Introduction}

Motion planning for Autonomous Vehicles (AVs) is particularly challenging in dense traffic, where limited space and complex interactions often lead to overly conservative behavior and unmet planning objectives. This stems from the AV's limited ability to account for the interactive responses of surrounding agents. Traditional decoupled prediction–planning methods often fall short in capturing how other drivers may adapt to the AV’s actions. For example, in merging scenarios, human drivers often anticipate yielding by other drivers, while AVs using non-interactive predictions may fail to act, resulting in inefficient or stalled maneuvers.

To address this issue, motion planning must incorporate interactive behavior models into decision-making. Advances in Machine Learning (ML) have enhanced trajectory prediction by capturing complex interactions from data. However, coupling prediction and decision-making remains challenging, especially with black-box, non-convex Neural Network (NN) models, which can be computationally intractable \cite{anjian_consistencyModel}. As a result, existing methods often rely on discrete motion primitives~\cite{nn-mpc, lanechange-rnn}, sacrificing optimality in continuous control, or analytical techniques that may hinder real-time performance~\cite{mpc-admm}.

Sampling-based optimization has gained popularity for controlling systems with black-box NN models~\cite{pets, drcc-mpc}. Among these, Model Predictive Path Integral (MPPI) control has shown strong performance in robotics~\cite{mppi-learned-dynamics, drcc-mpc}, aerial vehicles~\cite{pi-mppi, quadcopter-mppi}, and autonomous driving~\cite{mppi-learned-terrain, active-learning-mppi}, due to its sample efficiency, ability to handle nonlinear models, and lack of reliance on gradient information. Its rollout and evaluation phases are also highly parallelizable, enabling real-time control with thousands of samples.



\begin{figure}
\centering
\includegraphics[width=\linewidth, height=\linewidth, keepaspectratio]{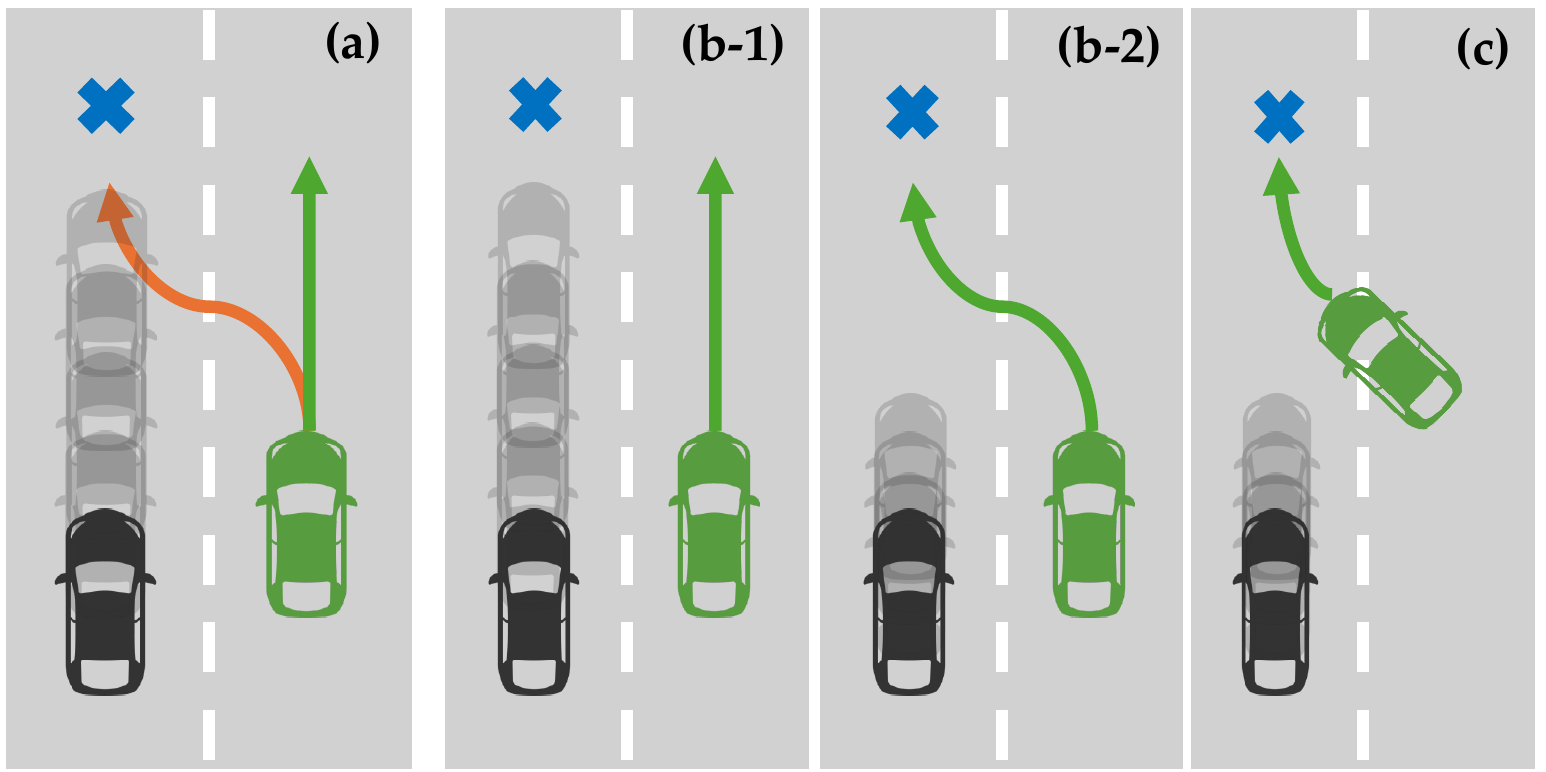}
\caption{\footnotesize Illustrative example of a merging scenario using interaction-aware MPPI. (a) In traditional decoupled prediction–planning frameworks with static predictions, the planner cannot model how surrounding vehicles react to the ego vehicle’s actions, resulting in inefficient behavior. In contrast, interaction-aware MPPI generates predictions conditioned on sampled control sequences. For instance, (b-1) shows nominal traffic flow when the ego continues straight, while (b-2) predicts the other vehicle yielding when the ego attempts to merge. (c) Using these reactive predictions, the ego vehicle performs a successful merge.}
\label{fig:main}
\end{figure}

We propose an interaction-aware motion planning framework that integrates a neural network-based interactive prediction model with an MPPI controller. The core idea is to predict the reactive behavior of surrounding vehicles based on each MPPI control sample. These predictions simulate the AV’s decision-making process. For instance, in a merging scenario, the prediction model may predict that a neighboring driver will yield when the AV attempts to merge, even if no gap initially exists. This allows the AV to assess the lane-change maneuver as safe and efficient. Additionally, the parallelization capabilities of MPPI enable real-time motion planning.

This work has two major contributions:
\begin{enumerate}
    \item We propose Interaction-Aware Neural Network-Enhanced Model Predictive Path Integral (IANN-MPPI), a real-time, fully parallelizable, interaction-aware trajectory planning framework that predicts the motion of surrounding agents based on MPPI control samples, enabling complex maneuvers. 
    

    \item We introduce a spline-based prior for MPPI sampling distributions to enhance IANN-MPPI in autonomous driving. This prior enables diverse sampling of control trajectories in structured lane environments, enabling efficient lane-changing behavior. 

\end{enumerate}

\section{Related Works}

\noindent\textbf{Interactive Motion Planning:}
Considering interaction behavior is critical for safe and efficient motion planning in multi-agent environments. Game-theoretic approaches model interactions as coupled optimization problems~\cite{inversegame-highway, stackelberg-ad}. While these approaches offer a clear mathematical framework and optimality, they often rely on assumptions about surrounding vehicles, like leader-follower~\cite{leader-follower-game} or shared objectives~\cite{game-motionplanning, potential-game}, limiting their real-world applicability.

Another line of work uses reinforcement learning (RL)~\cite{rl-motion-review} with neural network models, such as graph neural networks or attention mechanisms, to incorporate complex interactions into motion planning~\cite{gupta2022towards, constrained-rl-ad, attential-rl}. However, their lack of interpretability poses challenges for deployment in real-world autonomous driving, where thorough safety evaluations are crucial.

In autonomous driving, modular approaches for prediction and planning are popular due to their interpretability~\cite{avmotionplanning-review, slas, rcms, multifuture, fcp, bidirectional_overtaking}. A key challenge in interaction-aware planning within this framework is that prediction modules often rely on NN-based black-box models, making the optimization computationally demanding. Existing solutions typically sacrifice optimality by using discretized motion primitives~\cite{nn-mpc, lanechange-rnn} or compromise computational efficiency with analytical methods~\cite{mpc-admm}.

Our approach is similar to that of~\cite{ia-mppi}, using an NN prediction model to predict the motion of surrounding agents and apply MPPI control. However,~\cite{ia-mppi} uses the prediction model solely for high-level local goal prediction, with waypoints predicted using centralized MPPI, which requires known dynamics and cost functions for other agents. In contrast, our method directly uses NN predictions as predicted waypoints for surrounding vehicles, allowing us to account for low-level interaction behaviors learned from data.

\noindent\textbf{Model Predictive Path Integral (MPPI):} MPPI control~\cite{mppi} is a sampling-based receding horizon method grounded in information theory. Its sample efficiency, parallelizability, and independence from gradients make it well-suited for control systems integrated with neural networks, including learned dynamics~\cite{mppi-learned-dynamics}, cost functions~\cite{mpi-learned-cost, mppi-learned-value}, and terrain maps~\cite{mppi-learned-terrain}.

The design of the control sampling distribution greatly affects MPPI performance~\cite{mppi-covariance-design, cov-mppi}. Standard MPPI implementation utilizes a Gaussian distribution, that often fails to sample a diverse set of control samples or struggles when the optimal control sequence follows a multimodal distribution~\cite{stein-variational-mppi}. To address this, prior work has explored techniques like Stein variational gradient descent~\cite{stein-variational-mppi}, normalizing flows~\cite{normalizing-flow-mppi}, and auxiliary controllers~\cite{biased-mppi}. In this work, we adopt a spline motion prior with Biased-MPPI~\cite{biased-mppi} to encode lane information and promote diverse sampling. Unlike methods using discrete motion primitives~\cite{nn-mpc, lanechange-rnn}, MPPI’s continuous control enables flexible and effective planning around the spline prior. 

\section{Preliminaries and Problem Formulation}
\subsection{MPPI Review}~\label{subsec: MPPI_review}
We now provide an overview of MPPI framework~\cite{mppi, mppi-tro}. MPPI operates in three steps: \pg{Updated notation: U-control input, C- cost, $\mu$: mean - Check all the equations for correctness}

\begin{enumerate}
    \item \textbf{Sampling Control Trajectories:} To optimize control from time $t$, MPPI samples input sequences $U = (\mathbf{u}_{t}, \mathbf{u}_{t+1}, \dots, \mathbf{u}_{t+H-1})$ over a planning horizon $H$ from a predefined sampling distribution, often modeled as a Gaussian distribution, i.e., $\mathbf{u}_k \sim \mathcal{N}({\mu}_{k}, \Sigma)$, where, for $k\in\mathbb{N}$, $\mathbf{\mu}_{k}$ is the mean, and $\Sigma$ is the variance.

    \item \textbf{Cost Evaluation: } After sampling control inputs, MPPI rolls out a state trajectory $\mathbf{x}_{t:t+H} = (\mathbf{x}_{t}, \mathbf{x}_{t+t+1}, \dots, \mathbf{x}_{H})$ using the state dynamics model $\mathbf{x}_{k+1} = f(\mathbf{x}_{k},  \mathbf{u}_{k})$, and evaluates its cost $C$ using an objective function $\mathcal{J}$, i.e., $C(U) = \mathcal{J}(\mathbf{x}_{t:t+H})$.

    \item \textbf{Update Sampling Distribution:} MPPI updates the sampling distribution using importance-sampling weights, computed as:    \begin{multline}\label{eq:importance-weight}
    w(U) = \frac{1}{\eta} \exp \Bigg( -\frac{C(U)}{\lambda} + \frac{1}{2} \sum_{k=t}^{t+H-1} \mathbf{\mu}_{k}^\top \Sigma^{-1}\mathbf{\mu}_{k} \\ 
     - \sum_{k=t}^{t+H-1} \mathbf{\mu}_{k}^\top \Sigma^{-1}\mathbf{u}_k  \Bigg)
\end{multline}
where $\eta$ and $\lambda$ are hyperparameters. The optimal control mean is then estimated via Monte Carlo as:
\begin{align}\label{eq:monte-carlo-estimation}
    \mathbf{\mu}_t^* = \mathbb{E}_{\mathbf{u}_t \sim \mathcal{N}(\mathbf{\mu}_t, \Sigma)} [ w(U)\mathbf{u}_t ].
\end{align}
In practice, a time-shifted version of $\mathbf{\mu}_t^*$ is used as the sampling mean in the next time-step~\cite{mppi-tro}.




\end{enumerate}

\subsection{Problem Formulation}

Let $\mathbf{x}^i_t$ denote the state of vehicle $i \in \{1, \ldots, N^{veh}\}$ at time $t$, where $N^{veh}$ denotes the total number of surrounding vehicles. Surrounding vehicles are defined as those located in the same or adjacent lanes with the ego vehicle and within a specified distance $d$. We use the superscript $ego$ to denote the ego vehicle, e.g., $\mathbf{x}^{ego}_t$ represents the ego vehicle's state at time $t$. The global state vector of all the surrounding vehicles $\mathbf{x}_t^{veh}$ at time-step $t$ is given by:
\begin{align*}
    \mathbf{x}^{veh}_{t}  \triangleq [\mathbf{x}^1_{t},\ldots, \mathbf{x}^{N_{veh}}_{t}]^\top.
\end{align*}


Let $\Delta t$ denote the discrete time-step. We model the vehicle dynamics using the discrete-time nonlinear kinematic bicycle model~\cite{bicycle}: 
\begin{equation*}
    \begin{aligned}
       x_{t+1} &= x_{t} +  v_t\cos(\psi_t+\beta_t)\Delta t, \\
       y_{t+1} &= y_t + v_t\sin(\psi_t+\beta_t)\Delta t,\\
       \psi_{t+1} &= \psi_t + \frac{v_t}{l_r}\sin(\beta_t)\Delta t, \\
       v_{t+1} &= v_t + a_t\Delta t,\\
       \beta_t &= \tan^{-1}\left(\frac{l_r}{l_f+l_r}\tan(\delta_t)\right),
    \end{aligned}
\end{equation*}
where the state vector $\mathbf{x}^i_{t} = [x^i_{t}, y^i_{t}, \psi^i_{t}, v^i_{t}]$ represents [x-y coordinates, heading angle, speed] and the control input $\mathbf{u}^i_{t} = [\delta^i_{t}, a^i_{t}]$ represents [steering angle, acceleration] of the agent $i$ at time $t$.  The parameters $l_f$ and $l_r$ are the distances from the vehicle’s center to the front and rear axles, respectively.


We assume surrounding vehicles interact with each other and the ego vehicle, meaning their future states depend on the ego vehicle’s planned trajectory. We utilize this in our interaction-aware MPC formulation in~\eqref{subeq:MPC-prediction}.


Let ${0:t}$ denote time concatenation from $0$ to $t$. We formulate motion planning as a Model Predictive Control (MPC) problem with objective cost function $\mathcal{J}$  and MPC planning horizon $H$. The interaction-aware MPC problem is formulated as follows: 
\begin{problem}[Interaction-aware MPC] \label{problem:mpc formulation}
    \begin{subequations}
        \begin{align}
            &\min_{\mathbf{x}^{ego}_{t:t+H}, \mathbf{u}^{ego}_{t:t+H}} \mathcal{J} (\mathbf{x}^{ego}_{t:t+H}, \mathbf{u}^{ego}_{t:t+H}, \mathbf{x}^{veh}_{t:t+H_{pred}}),\\
  \text{s.t.}~~~&\mathbf{x}^{ego}_{t+1} = f(\mathbf{x}^{ego}_{t},\mathbf{u}^{ego}_{t})\label{subeq:MPC-bicycle},\\
            &\mathbf{x}^{veh}_{t+1:t+H_{pred}} =g(\mathbf{x}^{veh}_{0:t}, \mathbf{x}^{ego}_{0:t}, \mathbf{x}^{ego}_{t+1:t+H_{pred}}) \label{subeq:MPC-prediction} 
        \end{align}
    \end{subequations}
where $g(\cdot)$ denotes unknown interaction function and $H_{pred}$ is the prediction horizon. 
\end{problem}

The interactive prediction model in~\eqref{subeq:MPC-prediction} captures the influence of both past and planned ego trajectories on surrounding vehicle behavior. Note that the prediction horizon $H_{pred}$ may differ from the planning horizon $H$ for design flexibility.

\subsection{Planning Objective}
\begin{figure}
    \centering
    \includegraphics[width=0.7\linewidth, height=\linewidth, keepaspectratio]{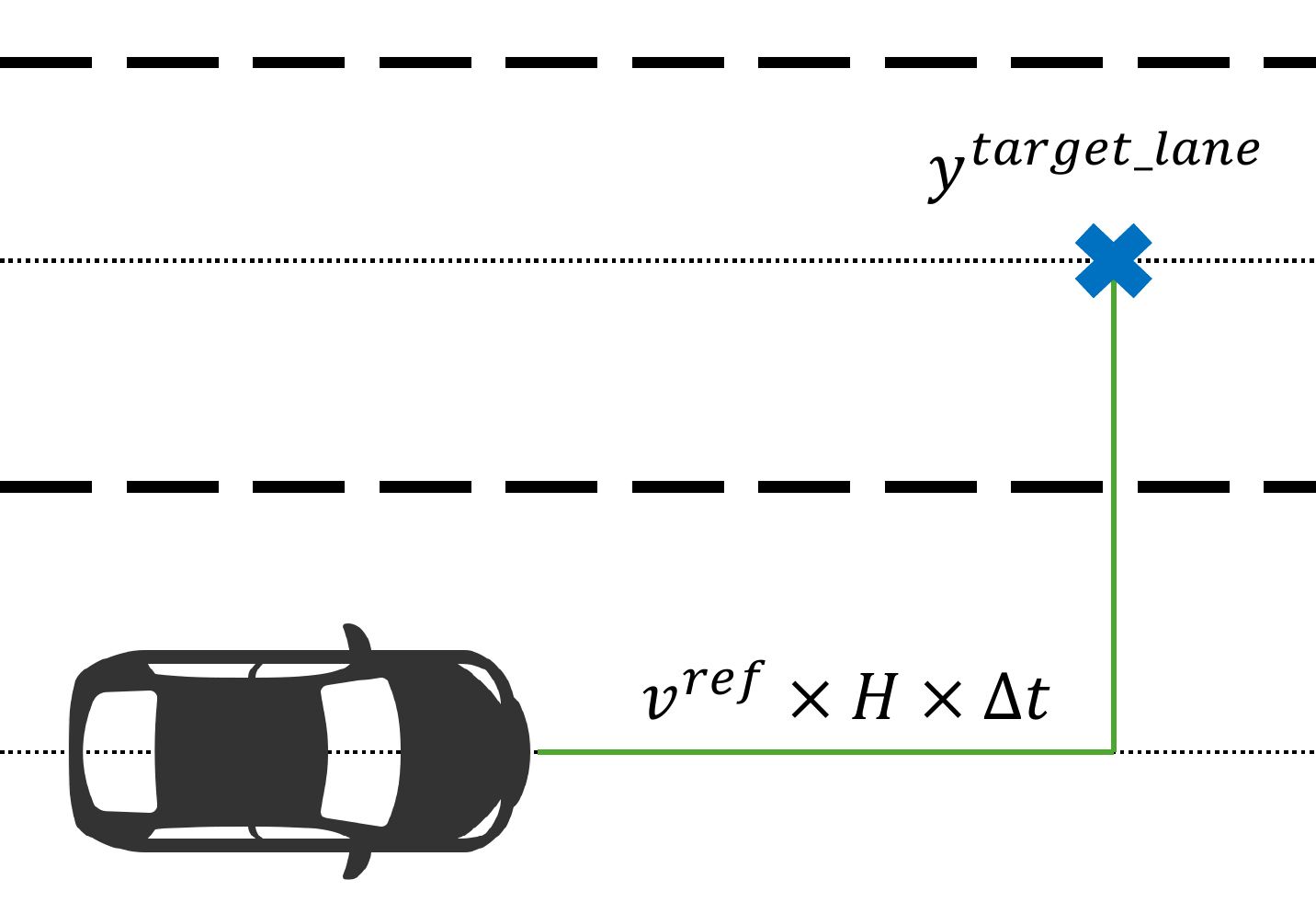}
    \caption{\footnotesize Illustrative figure for local goal design. The local goal directs the AV to maintain forward progress at a reference velocity while guiding it toward a merge into the target lane.}
    \label{fig:local-goal}
\end{figure}
We design the cost function $\mathcal{J}$ to achieve the planning objective (e.g., lane following or merging) while ensuring safety and driving comfort following~\cite{nn-mpc}. Specifically, $\mathcal{J}$ is defined as: 

\begin{align}\label{eq:objective}
    &\mathcal{J}(\mathbf{x}^{ego}_{t:t+H}, \mathbf{u}^{ego}_{t:t+H}, \mathbf{x}^{veh}_{t:t+H_{pred}}) = \notag\\
    &\sum_{k=t+1}^{t+H} \lambda^{goal} \mathmybb{1}(\|x^{ego}_{k} - x^{goal}_{k}\|^2 < \varepsilon ) && (\text{\hfill $x$ dir tracking}) \notag \\
    &+\sum_{k=t+1}^{t+H} \lambda^{goal} \mathmybb{1}(\|y^{ego}_{k} - y^{goal}_{k}\|^2 < \varepsilon ) && (\text{\hfill $y$ dir tracking} )\notag \\
    &+\sum_{k=t+1}^{t+H} \lambda^{l}\|y^{lane}_{k}-y^{ego}_{k}\|^2 && (\text{\hfill Lane centering}) \notag\\
    &+\sum_{k=t+1}^{t+H} \lambda^{v}\|v^{ref}_{k}-v^{ego}_{k}\|^2 && (\text{\hfill Velocity tracking} )\notag\\
    &+\sum_{k=t}^{t+H-1} \left(\lambda^\delta \|\delta_{k}\|^2 + \lambda^a\|a_{k}\|^2\right) && (\text{\hfill Control cost}) \notag\\
    &+\sum_{k=t}^{t+H-2} \lambda^j \|\delta_{k} - \delta_{k+1}\|^2 && \hspace*{-3mm}(\text{\hfill Steering rate cost}) \notag\\
    &+\sum_{k=t}^{t+H-2} \lambda^s\|a_{k}-a_{k+1}\|^2 && (\text{\hfill Jerk cost}) \notag\\
    &+\sum_{k=t+1}^{t+H} \lambda^b \log (1+ e^{-(y_k^{ego} - y^{boundary})^2}) && (\text{\hfill Road boundary}) \notag\\
    &+\sum_{k=t+1}^{t+H_{pred}} \sum_{i=1}^{N^{veh}} \lambda^\rho~\rho\left(\mathbf{x}^{i}_{k},\mathbf{x}^{ego}_{k} \right) && (\text{\hfill Safety risk}) 
\end{align}
\pg{Steering is $\delta$ and $\psi$ is the heading. Should it be heading cost instead of steering cost?}
where $\lambda^{goal}, \lambda^{l}, \lambda^{v}, \lambda^\delta, \lambda^a, \lambda^j, \lambda^s, \lambda^{b}, \lambda^\rho$ are positive hyperparameters, $(x^{ego}, y^{ego})$ is the Cartesian coordinate of the ego vehicle, $y^{lane}$ is the $y$ coordinate of the centerline of the ego lane, $y^{boundary}$ is the $y$ coordinate of the road boundary, $v^{ref}$ is the reference velocity, and $\rho(\cdot)$ is the risk measure to evaluate the safety risk with each vehicle configurations. $(x^{goal}, y^{goal})$ is the Cartesian coordinates of the local planning goal defined as follows (see Figure~\ref{fig:local-goal}):
\begin{align*}
    x^{goal} &= x^{ego} + v^{ref} \times H \times \Delta t \\
    y^{goal} &= y^{target\_lane}
\end{align*}

For the safety risk measure $\rho(\cdot)$, we adopt an ellipsoidal Gaussian risk model inspired by~\cite{rcms}, representing vehicle $i$ at time $t$ using a Gaussian distribution $\mathcal{N}(p^i_t, \Sigma^i_t)$ with mean $p^i_t = [x^i_t, y^i_t]$ and covariance $\Sigma^i_t$ defined as:
\small
\begin{align*}
    \Sigma^i_t = 
    \begin{bmatrix}
    \cos\psi^i_t & -\sin\psi^i_t \\
    \sin\psi^i_t & \cos\psi^i_t
    \end{bmatrix}
    \begin{bmatrix}
    \beta_LL & 0 \\
    0 & \beta_WW
    \end{bmatrix}
    \begin{bmatrix}
    \cos\psi^i_t & \sin\psi^i_t \\
    -\sin\psi^i_t & \cos\psi^i_t
    \end{bmatrix},
\end{align*}
\normalsize
where $L$ and $W$ denote the vehicle's length and width, while $\beta_L$ and $\beta_W$ are scaling factors. The Gaussian is centered at the vehicle's position, with a covariance ellipsoid reflecting its shape. The safety risk is then defined as the overlap between the ego vehicle's Gaussian and that of another vehicle.
\begin{align}
    \rho\left(\mathbf{x}^{i}_{t},\mathbf{x}^{ego}_{t} \right) = \iint \mathcal{N}(p^{ego}_t, \Sigma^{ego}_t) ~ \mathcal{N}(p^{i}_t, \Sigma^{i}_t) dxdy  \label{eq:gaussian_risk}
\end{align}

Since we know the integral of probability density function of Gaussian distribution, \eqref{eq:gaussian_risk} can be computed efficiently using analytical method~\cite{gaussian-risk}.


\section{Model Predictive Path Integral Control for Interaction Aware Planning}
\subsection{Interaction Aware Neural Network Trajectory Prediction}
\begin{figure}
    \centering
      \includegraphics[width=0.95\linewidth, height=\linewidth, keepaspectratio]{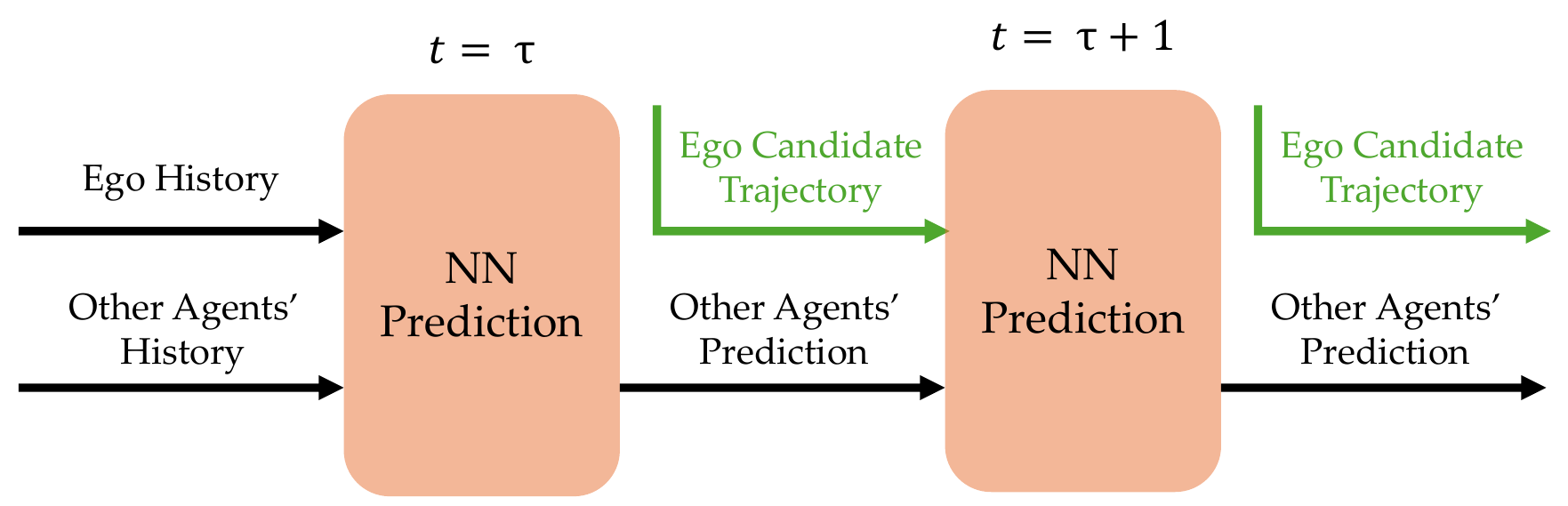}
    \caption{\footnotesize Rollout of the prediction model for generating ego-conditioned predictions. While the model captures interactions based on historical data, we obtain future-conditioned, interaction-aware predictions by performing multi-step rollouts and updating the ego vehicle’s history with the candidate trajectory. \ms{in the figure, shouldn't the history input to the NN consider something like $\tau-H,\tau-H+1,\ldots\tau-1$?  }
    \pg{I thing the time here is showing the time instance and is not with Other Agent's History. In that case, move the time at top of figure to avoid confusion?}}
    \label{fig:one-step-rollout}
\end{figure}

A key challenge in the interaction-aware MPC problem~\ref{problem:mpc formulation} is modeling the complex, unknown interactive behavior~\eqref{subeq:MPC-prediction}. To address this, we use a ML-based trajectory predictor $\phi$, trained to jointly predict the interactive future trajectories of the surrounding vehicles based on their trajectory history.


\begin{equation} \label{eq:nn-prediction}
    \mathbf{\hat{x}}^{veh}_{\tau+1} = \phi \left(\mathbf{x}^{veh}_{0:\tau}, \mathbf{x}^{ego}_{0:\tau} \right).
\end{equation}

Although the model predicts interactions based only on trajectory history, future interaction behaviors can be captured by rolling out the model using the ego vehicle’s candidate trajectory and one-step predictions of surrounding vehicles, as illustrated in Figure~\ref{fig:one-step-rollout}. The interactive prediction of the surrounding vehicles based on the ego's candidate trajectory is given by:
\begin{align}\label{eq:interactive_prediction}
    \mathbf{\hat{x}}^{veh}_{t+1} &= \phi \left(\mathbf{x}^{veh}_{0:t}, \mathbf{x}^{ego}_{0:t} \right) \notag \\
    \mathbf{\hat{x}}^{veh}_{t+2} &= \phi \left(\mathbf{x}^{veh}_{0:t}, \mathbf{x}^{ego}_{0:t}, \mathbf{\hat{x}}^{veh}_{t+1}, \mathbf{x}^{ego}_{t+1} \right) \notag\\
    &\dots \notag \\
    \mathbf{\hat{x}}^{veh}_{t+H_{pred}} &= \phi \left(\mathbf{x}^{veh}_{0:t}, \mathbf{x}^{ego}_{0:t}, \mathbf{\hat{x}}^{veh}_{t+1:t+H_{pred}-1}, \mathbf{x}^{ego}_{t+1:t+H_{pred}-1} \right) 
\end{align}

Our framework is agnostic to the specific prediction model, requiring only that it generates interaction-aware trajectories for surrounding vehicles. In this work, we use SGAN~\cite{sgan}, an RNN–GAN-based trajectory prediction model, which has been successfully integrated into autonomous driving motion planning~\cite{nn-mpc, lanechange-rnn, mpc-admm, particle-swarm}.


\noindent \textbf{\textit{Remark.}} Most NN-based prediction models capture interactive behaviors from past observations~\eqref{eq:nn-prediction}, not the ego's future trajectory. Conditioned predictions can still be generated by sequentially rolling out one-step inference with the ego's future trajectory, shown in~\eqref{eq:interactive_prediction}, at the cost of additional computation. Alternatively, joint prediction models~\cite{trajectron++, student-sgan} directly produce interaction-aware trajectories conditioned on the ego's future, eliminating iterative inference and improving efficiency.

\subsection{Joint Prediction on MPPI Control Samples}
Although we use a NN predictor in~\eqref{eq:nn-prediction} to approximate the unknown interaction model~\eqref{subeq:MPC-prediction}, solving Problem~\ref{problem:mpc formulation} remains difficult due to the non-convex, black-box nature of the model. To tackle this, we employ the MPPI framework and enable interactive planning by conditioning motion predictions on each sampled control trajectory.

As shown in Figure~\ref{fig:main}, when a control sample directs the ego vehicle to proceed straight, the prediction model assumes surrounding vehicles maintain nominal behavior. Conversely, if the sample involves merging, it anticipates the target-lane vehicle will yield. These conditional predictions enable our framework to simulate diverse interaction outcomes, supporting efficient and interaction-aware planning.

Following the MPPI procedure described in Sec.~\ref{subsec: MPPI_review}, we sample and roll out $K$ ego vehicle trajectories. For each, the corresponding interactive future trajectories of surrounding vehicles are generated using the NN predictor using~\eqref{eq:interactive_prediction}, and the cost is computed via~\eqref{eq:objective}.

\subsection{Integrating Spline Prior for Effective Merging}

To approach optimal solutions, MPPI must sample a diverse set of control sequences to capture varied interaction behaviors. However, standard MPPI often suffers from limited sample diversity due to its Gaussian sampling scheme, which tends to cluster around the mean and risks converging to suboptimal local minima.

To address this, we leverage discrete lane structures as prior information to guide ego behavior sampling. In autonomous driving, optimal trajectories typically involve maneuvers like lane-keeping or lane-changing. Instead of purely random sampling from a unimodal Gaussian, we encode this multi-lane structure into the sampling distribution, promoting diverse and context-aware trajectory generation.



\pg{How many splines are generated? Is there a single spline or multiple splines? In case of multiple splines, does M represent the sampled trajectories around single spline or all splines? Based on that change the next paragraph. Be clear about single/multiple splines} 

\pg{Needs a comment about what happens if a lane is missing. For example, no right lane. In that case only M samples are taken?}
For prior generation, we design polynomial spline curves for each neighboring lanes, leveraging their smoothness and computational efficiency~\cite{nn-mpc}. Specifically, we use cubic Hermite spline interpolation~\cite{cubic-hermite-spline} to generate waypoints for these maneuvers. To track the generated splines, we compute control sequences using a PID control for acceleration and the Stanley control~\cite{stanley-control} for steering. This yields a reference control sequence $U^{spline} = (\mathbf{u}^{spline}_t, \mathbf{u}^{spline}_{t+1}, \dots, \mathbf{u}^{spline}_{t+H-1})$. We then sample 
$M$ control sequences from Gaussians centered at the spline-based reference control sequence $U^{spline}$ for each neighboring lane, and remaining samples from the standard MPPI Gaussian distribution. Formally, the overall sampling distribution becomes:
\begin{align*}
    \mathbf{u}_t^k \sim 
    \begin{cases} 
    \mathcal{N} (\mathbf{\mu}^{spline\_left}_{t}, \Sigma^{spline}) & \text{if } k \leq M, \\
    \mathcal{N} (\mathbf{\mu}^{spline\_right}_{t}, \Sigma^{spline}) & \text{if } M < k \leq 2M, \\
    \mathcal{N} (\mathbf{\mu}_{t}, \Sigma) & \text{otherwise}.
    \end{cases}
\end{align*}


    


Integrating spline-based priors results in a non-Gaussian sampling distribution, rendering the original importance-sampling weights in Eq.~\eqref{eq:importance-weight} suboptimal. To address this, we adopt Biased-MPPI\cite{biased-mppi}, a variant of MPPI designed to handle arbitrary sampling distributions. It modifies the importance weights as follows:
\begin{align}\label{eq:biased-mppi-weight}
    w(U) = \frac{1}{\eta} \exp \left( -\frac{1}{\lambda} C(U) \right)
\end{align}
For a detailed explanation and derivation of~\eqref{eq:biased-mppi-weight}, we refer readers to\cite{biased-mppi}. The complete planning algorithm is summarized in Algorithm~\ref{alg:main}.

\begin{algorithm}
\caption{Interaction-aware MPPI}\label{alg:main}
\SetKwInput{KwGiven}{Given}
\KwGiven{$\mathcal{J}$ (cost function), $K$ (sampling population), $\lambda, \eta$ (MPPI parameter), $\Sigma, \Sigma^{spline}$ (MPPI sampling noise), $y^{target}$ (target lane) }
\KwIn{$\mathbf{x}_t^{ego}$ (current ego state) $\mathbf{x}_{0:t}^{veh}$ (history of surrounding vehicles state), $\mathbf{x}_{0:t}^{ego}$ (history of ego vehicles state) }
\BlankLine
\textbf{Step 1: Sample input sequence around spline prior motion}\\
$\text{Spline}_{left}, \text{Spline}_{right} \gets \text{CubicSpline}(\mathbf{x}_t^{ego}, y^{lane})$  \\
$\mu^{spline\_left}, \mu^{spline\_right} \gets \text{PathTracking}(\text{Spline}_{left}, \text{Spline}_{right})$ \\
\For{$k \gets 1$ to $M$}{ 
$ U^k \gets \mathcal{N}(\mu^{spline\_left}, \Sigma^{spline}),$ }
\For{$k \gets M+1$ to $2M$}{ 
$ U^k \gets \mathcal{N}(\mu^{spline\_right}, \Sigma^{spline}),$ }
\textbf{Step 2: Sample input sequence around previous solution}\\
\For{$k \gets 2M+1$ to $K$}{
\pg{Made it 2M+1}
$ U^k \gets \mathcal{N}(\mu, \Sigma), $} 

\BlankLine
\textbf{Step 3: Predict the reactive motions of surrounding vehicles}\\[1mm]
$ \mathbf{x}_{t+1:t+H}^{k, ego} \gets f(\mathbf{x}_t^{ego}, U^k) $ \\[1mm]
$ \hat{\mathbf{x}}_{t+1:t+H_{pred}}^{k, veh} \gets \phi(\mathbf{x}_{0:t}^{ego}, \mathbf{x}_{0:t}^{veh}, \mathbf{x}_{t+1:t+H}^{k, ego}) $\\[1mm]
$ \mathcal{J}^k \gets \mathcal{J}(\mathbf{x}_{t+1:t+H}^{k, ego}, \mathbf{u}_{t+1:t+H}^{k, ego}, \hat{\mathbf{x}}_{t+1:t+H_{pred}}^{k, veh}) $ \\

\BlankLine
\textbf{Step 4: Biased-MPPI optimization}\\
$C(U^k) = \mathcal{J}^k$ \\
$ w(U^k) \gets \frac{1}{\eta} \exp \left(-\frac{1}{\lambda} C(U^k) \right) $ \\
$ \mathbf{u}_t^* \gets \mathbb{E}[w(U)u_t] $ \\
$\mu \gets (\mathbf{u}_{t+1}^*, \mathbf{u}_{t+2}^*, \dots, \mathbf{u}_{t+H-1}^*)$

\Return $\mathbf{u}_t^*$
\end{algorithm}

\section{Experiments}

\subsection{Simulation Setup}
\begin{figure}
    \centering
    \includegraphics[width=0.7\linewidth]{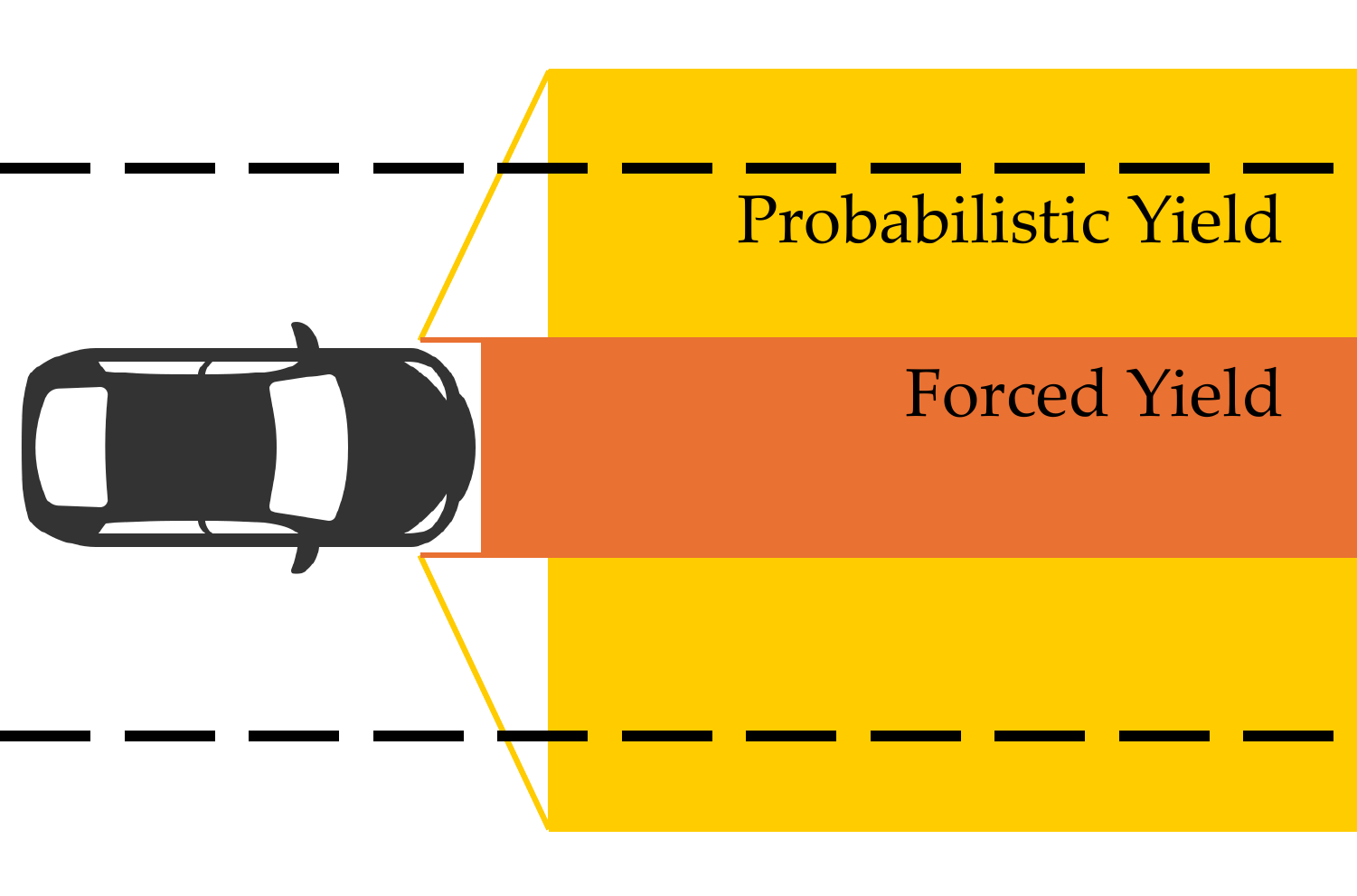}
    \caption{\footnotesize Forced yield zone and probabilistic yield zone of surrounding vehicles. When the ego vehicle enters the forced yield zone, the surrounding vehicle must yield to the ego vehicle. In contrast, in the probabilistic yield zone, the surrounding vehicle decides to yield or not with a predefined probability at each time step.}
    \label{fig:yield_zone}
\end{figure}

We evaluate IANN-MPPI in a dense-traffic merging scenario, where the ego vehicle aims to merge into a target lane occupied by multiple vehicles. Surrounding vehicles are simulated using three variations of the Intelligent Driver Model (IDM)\cite{IDM}:
\begin{enumerate}
    \item \textbf{Probabilistic IDM:}  Probabilistic IDM introduces a forced yield zone, where vehicles must yield if the ego is directly in their path, and a probabilistic yield zone, where yielding occurs with a predefined probability if the ego is nearby and attempting to merge (Figure~\ref{fig:yield_zone}).

    \item \textbf{Uncooperative IDM:} Uncooperative IDM excludes the probabilistic zone and only yields when the ego vehicle is directly ahead.

    \item \textbf{Cooperative IDM:} Cooperative IDM yields in both zones and maintains greater spacing from lead vehicles.
\end{enumerate}

\begin{table}
    \centering
    \caption{Simulation and MPPI parameters.} 
    \begin{tabular}{c c|c c}\toprule
        Params         & Values    & Params & Values \\ \midrule 
        $H$      & 17        &$N_{veh}$ & $5$  \\
        $H_{pred}$        & 8       & $\delta v^{ref}[m/s]$ & $\mathcal{U}(-1, 1)$  \\
        $K$      & 1500        &$\delta x_{init} [m] $           & $\mathcal{U}(-1, 1)$  \\
        $a_{min}, a_{max}[m/s^2]$ & $-0.5, 0.5 $        &$\delta v_{init} [m/s]$   & $\mathcal{U}(-1, 1)  $   \\
        $\psi_{min}, \psi_{max}[rad]$  & $-0.1, 0.1$  &$\Delta t$ & $0.3$       \\
        $\Sigma$     & $\text{diag}(0.1, 1\text{e-3})$         &$\lambda$                                    & $1$       \\
        $\Sigma^{spline}$       & $\text{diag}(0.1, 5\text{e-4})$       &$\eta$                                    & $1$       \\
    \bottomrule
    \end{tabular}
    \label{tab:simulation-params}
\end{table}

Additionally, for our ablation study, we evaluate three prediction models:
\begin{enumerate}
    \item \textbf{SGAN:} SGAN~\cite{sgan} is a neural network prediction model for IANN-MPPI, that captures interaction behaviors learned from data. SGAN utilize generative adversarial network (GAN) to generate future trajectory prediction based on history of ego and surrounding vehicles trajectories. We use Student-SGAN~\cite{student-sgan} which is trained with knowledge distillation for faster inference. We refer readers to~\cite{student-sgan} for training details.

    \item \textbf{Constant Velocity (CV):} CV serves as a non-interactive baseline, assuming vehicles maintain their current velocity over the prediction horizon. While simple, it performs well for short-term forecasting~\cite{constant-velocity} but lacks interaction modeling.

    \item \textbf{IDM:} IDM models interactions without neural networks, assuming surrounding vehicles always yield to the ego vehicle. Although it captures basic interactions, it cannot represent non-cooperative behaviors, resulting in overly aggressive behavior.
\end{enumerate}

In our Monte-Carlo simulations, we randomize the initial positions, velocities of surrounding vehicles, and the reference velocity. We test in various velocity scenarios by setting default $v^{ref}=2.5$ and randomizing $v^{ref}$ and $v_{init}$ with uniform noise $\mathcal{U}(-1, 1)$. Additional simulation and MPPI parameters are listed in Table~\ref{tab:simulation-params}. \jd{v=2 and 3 both sound like low speed experiements. maybe just averaging out the results might be better. other option is to claim differnt gap sizes - very tight to medium size gaps base don the intial spacing changes}


\subsection{Results}

\begin{table*}[htb!]
\centering
\caption{Simulation Results}
\begin{tabular}{cccccccc}
    \toprule
    Behavior Model & Prediction Model & Success($\%$)~$\uparrow$ & Collision(\%)~$\downarrow$ & Planning Cost~$\downarrow$ & Merge Time~$\downarrow$ & Acceleration~$\downarrow$ & Steering Rate~$\downarrow$ \\[2mm]
    \hdashline \\[-2mm]
    \multirow{3}{*}{\shortstack{Probabilistic\\IDM}} & SGAN & $\color{mycitecolor}\mathbf{67.5}$ & $0.0$ & $6.09 \pm 2.35$ & $24.35\pm16.13$ & $2.27 \pm 0.40$ & $0.35 \pm 0.08$ \\
      & CV & $\color{red}45.0$ & $0.0$ & $7.08\pm2.22$ & $32.94\pm15.26$ & $2.32\pm0.24$ & $0.31\pm0.06$ \\
     & IDM & $\color{mycitecolor}\mathbf{87.5}$ & $0.0$ & $5.03\pm2.08$ & $13.63\pm12.48$ & $1.76\pm0.37$ & $0.33\pm0.06$ \\[2mm]
    \hdashline \\[-2mm]
      \multirow{3}{*}{\shortstack{Uncooperative\\IDM}} & SGAN & $32.5$ & $\color{mycitecolor}\mathbf{0.0}$ & $7.76\pm1.97$ & $34.74\pm14.33$ & $2.35\pm0.40$ & $0.33\pm0.11$ \\
      & CV & $10.0$ & $\color{mycitecolor}\mathbf{0.0}$ & $8.58\pm1.55$ & $41.14\pm11.01$ & $2.46\pm0.33$ & $0.31\pm0.12$ \\
     & IDM & $40.0$ & $\color{red}32.5$ & $5.85\pm2.29$ & $8.69\pm15.57$ & $2.23\pm0.35$ & $0.45\pm0.14$ \\[2mm]
    \hdashline \\[-2mm]
     \multirow{3}{*}{\shortstack{Cooperative\\IDM}} & SGAN & $100.0$ & $0.0$ & $4.47\pm1.62$ & $8.10\pm1.89$ & $1.82\pm0.22$ & $0.38\pm0.00$ \\
      & CV & $100.0$ & $0.0$ & $4.59\pm1.72$ & $8.90\pm2.14$ & $1.85\pm0.26$ & $0.38\pm0.08$ \\
     & IDM & $100.0$ & $0.0$ & $4.40\pm1.48$ & $7.05\pm1.72$ & $1.62\pm0.21$ & $0.35\pm0.07$ \\
     \bottomrule
\end{tabular}
\label{table:result_summary}
\end{table*}

\begin{figure*}
    \centering
    \includegraphics[width=1.0\linewidth]{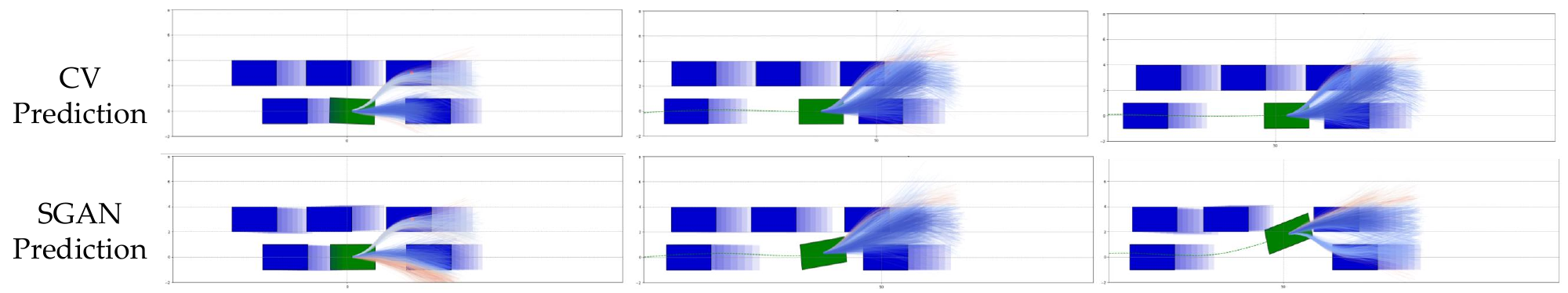}
    \caption{\footnotesize Comparison of MPPI with non-interaction-aware CV prediction (top) and our IANN-MPPI with SGAN prediction (bottom). SGAN predicts that the surrounding vehicle will slow down (middle column), enabling the ego vehicle to merge via nudging (final column). In contrast, CV fails to capture this interaction, leaving the ego vehicle in its source lane.}
    \label{fig:sgan_vs_cv_prediction}
\end{figure*}

We perform $40$ Monte-Carlo simulations per scenario and summarize the results in Table~\ref{table:result_summary}. Metrics include the success rate of merging, collision rate (based on a three-circle vehicle model~\cite{lanechange-rnn, particle-swarm}), planning cost (objective in~\eqref{eq:objective} excluding the risk term $\rho$), and merge time—the time taken by the ego vehicle to complete the merge. Additionally, we report absolute acceleration and steering rate as indicators of ride comfort. Planning cost and merge time jointly reflect the efficiency of the merging behavior.


Our results show that IANN-MPPI enables efficient and successful merging. As an initial validation, we evaluate each prediction model under a cooperative IDM scenario as a sanity check. In this scenario, where surrounding vehicles maintain larger gaps from their respective lead vehicles, both interaction-aware and non-interaction-aware prediction models are able to successfully complete the merge. However, in the more challenging probabilistic IDM scenario, IANN-MPPI achieves higher success rates compared to the non-interaction-aware CV baseline. As illustrated in Figure~\ref{fig:sgan_vs_cv_prediction}, the CV model assumes constant velocity for surrounding vehicles, failing to anticipate yielding behavior. In contrast, the SGAN-based prediction predicts the surrounding vehicle slowing down in response to the ego vehicle’s merging attempt, enabling effective nudging and higher success. The success rate increases with IDM-based prediction, which assumes surrounding vehicles consistently yield, thereby encouraging assertive merging attempts.

\begin{figure*}
    \centering
    \includegraphics[width=1.0\linewidth]{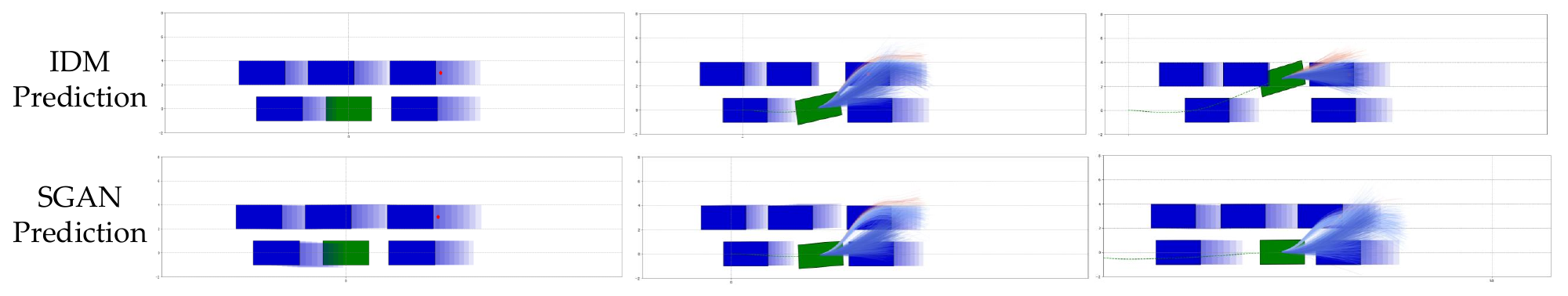}
    \caption{\footnotesize Comparison of IANN-MPPI with IDM prediction (top) and SGAN prediction (bottom) in uncooperative scenario. IDM assumes the surrounding vehicle will yield, prompting the ego vehicle to merge (middle column). When this assumption fails and the surrounding vehicle does not yield, a collision occurs (final column). In contrast, SGAN initially predicts yielding but adapts to the lack of cooperation, prompting the ego vehicle to return to its lane and avoid collision.}
    \label{fig:sgan_vs_idm_prediction}
\end{figure*}

However, IDM’s assumption that other vehicles will always yield can lead to overly aggressive merging, resulting in collisions when yielding does not occur. This is highlighted in high collision rate of IDM prediction model paired with uncooperative IDM scenario. As shown in Figure~\ref{fig:sgan_vs_idm_prediction}, the IDM model incorrectly predicts yielding behavior, causing a collision. In comparison, the SGAN model accurately infers that the surrounding vehicle will not yield and prompts the ego vehicle to abort the merge and return to its lane. This highlights the value of NN-based predictors in capturing nuanced interactions beyond simple heuristic methods.

\subsection{Spline Prior enhances Efficient Merging}

We observe that incorporating the spline prior improves both efficiency and success rates. Table \ref{tab:spline-curve} shows that the success rate improves from 0.75 to 0.8 and the merge time drops significantly (by ~10 seconds). The planning cost also decreases substantially, indicating that the spline prior helps guide the MPPI search toward more optimal and context-aware maneuvers.
\begin{table}[ht]
    \centering
    \begin{tabular}{cccc}
    \toprule
        MPPI & Success $\uparrow$ & Merge Time $\downarrow$ & Planning Cost $\downarrow$ \\[2mm]
        \hdashline \\[-2mm]
        with spline prior & $0.8$ & $21.40\pm11.33$ & $5.90 \pm 2.48$ \\
        without spline prior & $0.75$ & $31.83\pm8.23$ & $9.49\pm3.47$ \\
    \bottomrule
    \end{tabular}
    \caption{Effects of Spline-prior}
    \label{tab:spline-curve}
\end{table}

\subsection{Computation Time}
A key advantage of MPPI is its computational efficiency, enabled by GPU-based parallelization and sample efficiency. In our simulation setup with a planning horizon of $H=17$ (5.1s) and prediction horizon of $H_{pred}=8$ (2.4s), we use $K=1500$ samples per iteration. All evaluations were conducted on a Linux system with an Intel i9-12900HK CPU and an NVIDIA GeForce RTX 3080 Ti Mobile GPU. MPPI and NN inference were parallelized using PyTorch with GPU acceleration.
\jd{edit comp time and differentite with teacher network}

Across 20 runs, the average and standard deviation of computation time per planning step was $0.10\pm0.02$ seconds. Real-time computation was possible with Student-SGAN~\cite{student-sgan}, which enables faster inference of SGAN using knowledge distillation. For comparison, computation time with original SGAN is $0.34 \pm 0.02$ and constant velocity (CV) predictor is $0.026 \pm 0.003$ seconds.  

\jd{we need to tone down some claims since IDM baseline works better in many cases. highlight perfect prediciton of IDM, potential upper bound. highlight usefulness of our proposal and tradeoffs. 
"IANN-MPPI is model-agnostic and can integrate future prediction modules or uncertainty-aware frameworks, making it suitable for broader deployment" . 
"Although SGAN improves interactivity modeling, it increases computational load. This trade-off must be managed depending on hardware availability and safety-critical latency constraints."}

\section{Conclusions}
In this paper, we proposed IANN-MPPI, which enables interaction-aware trajectory planning by integrating neural network-based prediction with Model Predictive Path Integral (MPPI) control. IANN-MPPI captures surrounding agents' reactive behaviors conditioned on MPPI control samples, enabling cost evaluation that accounts for their responses to the ego vehicle. Additionally, spline-based priors are incorporated in the sampling process to facilitate effective merging maneuvers. In simulation studies, IANN-MPPI achieves successful merging in dense traffic, proactively nudging surrounding vehicles while avoiding excessive aggressiveness of purely cooperative models. Furthermore, real-time performance is achieved through GPU parallelization.

The main limitation of this work is to disregard the prediction uncertainty. Incorporating uncertainty-aware trajectory predictors~\cite{trajectron++, mtr} or extending the MPPI framework with prediction distributions~\cite{ensemble-mppi} may improve safety and practicality. Furthermore, while the NN-based predictor improves interaction modeling, it may be biased toward cooperation if trained on datasets lacking non-cooperative scenarios like collisions, highlighting the need for careful data curation. Lastly, incorporating an NN prediction module increases computational overhead. This trade-off must be balanced considering hardware resources and latency constraints. Future work may explore extending the proposed framework to other interaction-sensitive domains such as crowd navigation.

\bibliographystyle{unsrt}
\bibliography{refs}






\end{document}